# Self-Supervised Bernoulli Autoencoders for Semi-Supervised Hashing

Ricardo Ñanculef, Francisco Mena
Department of Informatics
Federico Santa María Technical University
8940572 Santiago, Chile
Emails: {ricardo.nanculef, francisco.menat}@usm.cl

Antonio Macaluso, Stefano Lodi, Claudio Sartori
Department of Computer Science and Engineering
University of Bologna
40136 Bologna BO, Italy
Emails: {antonio.macaluso2, stefano.lodi, claudio.sartori}@unibo.it

*Abstract*—Semantic hashing is an emerging technique for large-scale similarity search based on representing high-dimensional data using similarity-preserving binary codes used for efficient indexing and search. It has recently been shown that variational autoencoders, with Bernoulli latent representations parametrized by neural nets, can be successfully trained to learn such codes in supervised and unsupervised scenarios, improving on more traditional methods thanks to their ability to handle the binary constraints architecturally. However, the scenario where labels are scarce has not been studied yet.

This paper investigates the robustness of hashing methods based on variational autoencoders to the lack of supervision, focusing on two semi-supervised approaches currently in use. The first augments the variational autoencoder's training objective to jointly model the distribution over the data and the class labels. The second approach exploits the annotations to define an additional pairwise loss that enforces consistency between the similarity in the code (Hamming) space and the similarity in the label space. Our experiments show that both methods can significantly increase the hash codes' quality. The pairwise approach can exhibit an advantage when the number of labelled points is large. However, we found that this method degrades quickly and loses its advantage when labelled samples decrease. To circumvent this problem, we propose a novel supervision method in which the model uses its label distribution predictions to implement the pairwise objective. Compared to the best baseline, this procedure yields similar performance in fully supervised settings but improves significantly the results when labelled data is scarce. Our code is made publicly available at https://github.com/amacaluso/SSB-VAE.

## I. INTRODUCTION

Given a dataset $D = \{x^{(1)}, x^{(2)}, \ldots, x^{(N)}\}$, with $x^{(\ell)} \in \mathbb{X} \ \forall \ell \in [N]$, *similarity search* is the problem of finding the elements of $D$ that are *similar* to a query object $q \in \mathbb{X}$, not necessarily in $D$. This is a fundamental task in computer science, lying at the foundation of many algorithms for pattern recognition. The rapid increase in the amount of high-dimensional data such as images, audio and text, has increased the interest for this type of search in the last years and raised the need for methods that can approach the task with reduced processing time and memory footprint.

If $\mathbb{X}$ is equipped with a similarity function $s : \mathbb{X} \times \mathbb{X} \to \mathbb{R}^1$ and $n$ is small, a simple approach to solve this problem is a *linear scan*: compare $q$ with all the elements in $D$

[1] the greater the value of $s$, the more similar are the objects.

and return $x^{(\ell)}$ if $s(x^{(\ell)}, q)$ is greater than some threshold $\theta$. The value of $\theta$ can be given in advance, computed to return exactly $k$ results or, more often, chosen to maximize information retrieval metrics such as precision and recall [1]. If $\mathbb{X} \subset \mathbb{R}^d$, with small $d$, tree-based indexing methods such as KD-trees have been traditionally used to perform efficient scans when $N$ is large. Unfortunately, if also $d$ is large, the computational performance of these data structures quickly degrades. Besides, if the similarity function $s$ determining the relevant results is not perfectly known, these methods cannot be used.

Semantic hashing deals with similarity search by *learning* a similarity-preserving hash function $h(x) \in \{0, 1\}^B$ that maps similar data to nearby positions in a hash table, preventing at the same time undesirable collisions. Items similar to a query $q$ can be easily found by just accessing all the cells of the table that differ a few bits from $h(q)$. As binary codes are storage-efficient, these operations can be performed in main memory even for very large datasets.

Although early hashing algorithms were randomized methods devised to preserve specific and well-known similarity functions (e.g. cosine) [8], it was soon realized that methods based on machine learning could significantly reduce the number ($B$) of bits required to preserve similarity by exploiting the fact that real data is often not uniformly distributed in $\mathbb{X}$. One of the first methods of this type [19] used a deep probabilistic model to learn a *manifold* underlying the data distribution. Unfortunately, training this model was hard in practice and perhaps for this reason, most subsequent research on hashing preferred to adopt more shallow architectures or, slightly later, deterministic neural network models that were easier to train. Shallow algorithms that flowered in this period include Spectral Hashing [25], Iterative Quantization [6], Kernel Locality Sensitive Hashing [12] and LDA-Hash [21]. Popular examples of non-probabilistic deep learning methods include the Binary Autoencoder of [2], UH-BDNN [5], Deep Hashing [15], most of them based on deterministic autoencoders augmented with constraints.

In the last years, machine learning has seen a renewed interest in probabilistic graphical models parametrized by neural nets. A hallmark of this approach is the ability to back-propagate gradients through stochastic layers with low variance [10, 9], which has permitted to scale these models to very large datasets and improve performance in many tasks. In particular, stochastic models taking advantage of variational autoencoders have shown to systematically outperform more traditional hashing algorithms [3]. It has been shown that this advantage can be further improved using Bernoulli latent representations that naturally encode the binary constraint underlying hash codes and thus reduce the quantization error arising from continuous representations [17]. Other recent contributions have shown that these models can be easily extended to leverage supervision, that is labels conveying information about the semantic content of the items to be indexed. Two different supervision schemes have been proposed. One of them, often termed *pointwise* supervision in earlier literature, augments the training objective to predict the label distribution of a training pattern [3]. The other method, often termed *pairwise* supervision in previous art, exploits the labels to define an additional objective in which pairs of items with the same label are required to have similar hash codes and pairs of items with different labels are enforced to have different hash codes. This approach yields state-of-the-art performance in [4] assuming that the labels of all the training examples are known.

As in many real-world tasks, obtaining labelled data is difficult and time-consuming, understanding the efficacy of the current methods to exploit annotations when they are scarce is a matter of significant importance in practice. In this paper, we show that even if using the ground-truth labels to introduce *pairwise* constraints can yield advantages when the number of labelled samples is large, this approach degrades quickly as the level of supervision decreases. We found that in many cases, it actually looses its advantage with respect to a model employing *pointwise* supervision only. To overcome this limitation, we propose to equip the variational autoencoder with a novel mechanism to exploit annotations in which the label distributions required for the pairwise loss are replaced by the model's own predictions. Ground-truth labels are still used to supervise the estimation of the label distributions but the pairwise constraints ask now for a consistency between the codes and the model's own beliefs about the class of the patterns. Experiments on text and image retrieval tasks show that this procedure is competitive to the best baseline in scenarios of label abundance, but it is more effective in scenarios of label scarcity.

## II. Related Work

The problem of representing high-dimensional data using binary codes that preserve their semantic content and support efficient indexing has been extensively studied in the literature. Traditionally, a first distinction is made between data dependent and data independent methods. Data independent methods are randomized techniques devised to preserve specific similarity functions (e.g. cosine) or distance metrics (e.g. Euclidean) [8]. They have good theoretical grounds, but usually they require long hash codes to achieve satisfactory performance. Data dependent methods can often obtain more effective and compact hash codes by learning the underlying structure of the dataset [25, 19]. Many techniques to achieve this goal have been studied in the last years, including unsupervised, supervised, and semi-supervised approaches.

Unsupervised methods rely purely on the properties of the points to be indexed. For instance, *Iterative Quantization* (ITQ) [6] computes the codes by applying PCA followed by a rotation that minimizes the quantization error arising from thresholding. *Spectral Hashing* (SpH) [25] poses hashing as the problem of partitioning a graph that encodes information about the geometry of the dataset. Recently, more flexible unsupervised models based on autoencoders have been proposed. In [2], a classic (deterministic) autoencoder is trained for hashing by minimizing the reconstruction error with an explicit constraint to handle the quantization error. The constraint prevents the model to be trained with back-propagation and a mixed integer programming solver needs to be applied. The method in [5] employs an deeper architecture for the encoder but keeps the architecture of the decoder. Binary representations are enforced in the vein of [2], using constraints. In spite of their computational complexity, experimental results of [2] and [5] are encouraging and suggest that hashing based on autoencoders can outperform other deep learning methods such as [16]. Our work is more related to the methods in [3], [17] and [4], which make use of deep variational autoencoders for hashing. Unlike classic autoencoders that learn a one-to-one map between the input space and the Hamming space, variational autoencoders learn the most likely region of the code space where an input pattern should be allocated. In order to reconstruct data from the latent representation, the model is constrained to place similar data in a similar region of the code space, reproducing in the training phase the mechanism that will be used later for search. [3] showed that this fundamental difference between classic and variational autoencoders is relevant for hashing and yields to significantly better results. Later, [17] demonstrated that the use of Bernoulli instead of Gaussian latent variables helps to reduce the quantization loss arising from the use of continuous representations. This idea is also used in [4] and extended to incorporate supervision.

Supervised hashing algorithms leverage information about the semantics of the items in the dataset to improve the hash codes. Pointwise methods such as [24] use labels or tags to implicitly enforce a consistency between the codes and the annotations. Pairwise methods assume that *pairs* of objects have been annotated as similar or dissimilar and attempt to explicitly preserve these similarities in the formulation. In [16] pairwise similarity relations are derived from class labels and integrated into the training objective of a neural net. Deep methods that leverage information from triplets or lists have also been proposed [13]. A different form of supervision that is often combined with deep learning techniques is *self-taught hashing* [28]. In this approach, a classic (often shallow) method is first used to find hash codes for a set of training examples. Then, a supervised (often deep) model is used to predict the codes of unseen data. A limitation of this approach is the sensitivity of the first step to the features used to represent the data, which may require to iterate the two steps.

Hashing methods tailored to semi-supervised scenarios, in which labelled and unlabelled samples are available, have started to be studied in the last years. They integrate supervised and unsupervised learning mechanisms. The method in [5] extends the objective function of a traditional (unsupervised) auto-encoder with an term based on pairwise supervision. [23] presents various methods based on linear projections, which combine pairwise supervision with an unsupervised learning goal inspired in information theory (max entropy). In [20], pointwise and pairwise supervision schemes and combined with spectral methods [25]. [18] uses pairwise and triplet-wise supervision to extend ITQ, a linear method for unsupervised hashing [6]. Building on the idea of self-training [22], a classic approach for dealing with partially labelled datasets, [27] has recently explored an iterative method in which label representations and hash codes are learned together in the model. By predicting the labels of data without annotations, the labelled dataset can be expanded and the model retrained. While fairly successful, this hashing model requires iterative re-training. Our method differs from this approach also in that it incorporates an explicit unsupervised learning mechanism - the supervised one being complementary. Other recent methods termed self-trained or self-supervised in the literature use actually a different approach. The method in [14] is based on learning label encodings that substitute standard one-hot vectors. The method in [26] is actually an application of self-taught hashing to cross-modality hashing in which the first stage exploits pairwise supervision. In computer vision, self-supervised hashing methods often uses domain information that cannot be easily generalized to other tasks. For instance [29] uses the different frames of a video to build similar pairs of examples and randomly picked frames to form dissimilar pairs. Image rotations or image patching have also been employed in other works.

Besides being able to learn compact and effective hash codes in a principled unsupervised way, variational autoencoders can be easily extended to exploit annotations. The seminal method in [3] has shown indeed that by training the model to learn both the data distribution and the label distribution, one can dramatically increase the efficacy of the hash codes in similarity search tasks. Building on this idea, [4] proposed to further extend the objective function of the model by using pairwise supervision. This approach yields state-of-the-art performance in [4] assuming that the labels of all the training examples are known. Our work extends these recent studies by considering a semi-supervised scenario in which a small set of instances have been annotated with class labels. We show that in this setting, the advantage of the pairwise method can significantly degrade when the number of annotated samples is small. Up to the best of our knowledge, the method we propose to deal with this issue has not been previously explored. Certainly, it can be connected with co-supervised methods [22] in which two or more learners iteratively teach each other to substitute the lack of annotations. These methods often used different and conditionally independent feature representations of the data. Our method does not use multiple training stages, resorts on a single feature representation, is used inside the same model and is specifically tailored to hashing with variational autoencoders.

## III. METHODS

### A. Generative Model

As in related works [17], we pose hashing as an inference problem, where the objective is to learn a probability distribution $q_\phi(\boldsymbol{b}|\boldsymbol{x})$ of the code $\boldsymbol{b} \in \{0,1\}^B$ corresponding to an input pattern $\boldsymbol{x}$. This framework is based on a generative process involving two steps: (i) choose an entry of the hash table according to some probability distribution $p_\theta(\boldsymbol{b})$, and (ii) sample an observation $\boldsymbol{x}$ indexed by that address according to a conditional distribution $p_\theta(\boldsymbol{x}|\boldsymbol{b})$. The parameters of this random process are learnt in such a way that it approximates the real data distribution.

### B. Bernoulli Autoencoders

Following [10], the distribution $q_\phi(\boldsymbol{b}|\boldsymbol{x})$ is called *the encoder*, and the distribution $p_\theta(\boldsymbol{x}|\boldsymbol{b})$ *the decoder*. In the original construction, $q_\phi(\boldsymbol{b}|\boldsymbol{x})$ is chosen to be a Gaussian $\mathcal{N}(\mu_\phi(\boldsymbol{x}), \sigma_\phi^2(\boldsymbol{x}))$ and binary codes are obtained by thresholding $\mu_\phi(\boldsymbol{x})$ around its empirical median [3]. In Bernoulli variational autoencoders (B-VAEs) in contrast, the encoder is chosen to be a multi-variate Bernoulli $\text{Ber}(\alpha(\boldsymbol{x}))$ with activation probabilities $\alpha(\boldsymbol{x})$. This choice permits to handle the binary constraint underlying hashing in an architectural way, creating an inductive bias that can significantly reduce the quantization loss incurred from thresholding Gaussian representations [17].

### C. Parametrization by Neural Nets

To learn flexible non-linear mappings, the activation probabilities of the encoder $\alpha(\boldsymbol{x})$ can be represented using a neural net $f(\boldsymbol{x}; \phi)$. The architecture of this model is chosen according to the dataset. In the simplest case, it is obtained as the composition of $L$ fully-connected layers $f_1 \circ \ldots f_{L-1} \circ f_L$ where $f_1 : \mathbb{X} \to \mathbb{R}^{n_1}$ accommodates the input data (a feature vector) and $f_L : \mathbb{R}^{n_L} \to [0,1]^B$ produces the activation probabilities. The latter is usually obtained by using a layer of independent sigmoid neurons [7].

The architecture for the decoder depends also on the application. In regression problems with real data, $p_\theta(\boldsymbol{x}|\boldsymbol{b})$ is often implemented using a Gaussian $\mathcal{N}(\mu_\theta(\boldsymbol{x}), \sigma^2)$ where $\mu_\theta((\boldsymbol{x})$ is predicted using a neural net $g(\boldsymbol{b}; \theta)$ with linear output layer. In text and image retrieval applications it is more common to represent the data using normalized features $\boldsymbol{x}_i \in [0,1]$ (word frequencies or pixels). In this case $g(\boldsymbol{b}; \theta)$ can be implemented using a net $g_1 \circ \ldots g_{L'-1} \circ g_{L'}$ that ends with a layer of sigmoid neurons (other layers can use other activations, e.g. ReLU).

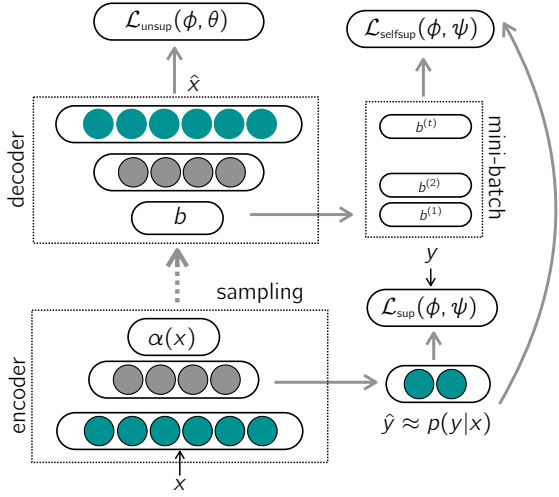

Fig. 1. Sketch of the architectures studied in this paper.

## D. Unsupervised Training

As illustrated in Fig.1, the composition of the encoder $p_\theta(\boldsymbol{x}|\boldsymbol{b})$ and the decoder $q_\phi(\boldsymbol{b}|\boldsymbol{x})$ leads to a stochastic autoencoder. This model can be trained without supervision using variational methods [10]. If $S = \{\boldsymbol{x}^{(1)}, \boldsymbol{x}^{(2)}, \ldots, \boldsymbol{x}^{(n)}\}$ denotes the set of training examples, the negative log-likelihood corresponding to a single data point $\boldsymbol{x}^{(\ell)} \in S$, can be upper bounded by the following loss function [17]:

$$\mathcal{L}_{\text{unsup}}^{(\ell)} = \mathbb{E}_{q_\phi(\boldsymbol{b}|\boldsymbol{x}^{(\ell)})}\left[-\log p_\theta(\boldsymbol{x}^{(\ell)}, \boldsymbol{b}) + \log q_\phi(\boldsymbol{b}|\boldsymbol{x}^{(\ell)})\right] \quad (1)$$
$$= \underbrace{\mathbb{E}_{q_\phi(\boldsymbol{b}|\boldsymbol{x}^{(\ell)})}\left[-\log p_\theta(\boldsymbol{x}^{(\ell)}|\boldsymbol{b})\right]}_{\mathcal{L}_1} + \underbrace{\lambda\, D_{\text{KL}}\left(q_\phi(\boldsymbol{b}|\boldsymbol{x}^{(\ell)}) \| p_\theta(\boldsymbol{b})\right)}_{\mathcal{L}_2}.$$

The term $\mathcal{L}_1$ measures the expected error in the reconstruction of $\boldsymbol{x}$ from the hash code $\boldsymbol{b}$. For instance, if the decoder is Gaussian, $\mathcal{L}_1$ is proportional to the squared loss $\|\hat{\boldsymbol{x}} - \boldsymbol{x}\|^2$ between the decoder's output $\hat{\boldsymbol{x}}$ and the original observation $\boldsymbol{x}$. The term $\mathcal{L}_2$ on the other hand, measures the Kullback-Leibler divergence between the distribution learnt by the encoder $q_\phi(\boldsymbol{b}|\boldsymbol{x})$ and some prior $p_\theta(\boldsymbol{b})$. For hashing applications with Bernoulli autoencoders, this can be chosen as $p_\theta(\boldsymbol{b}_i) = \text{Ber}(0.5)\ \forall i \in [B]$, which expresses a preference for balanced hash tables. With this choice, $\mathcal{L}_2$ can be computed analytically [17].

The main complexity of optimizing (1) using standard techniques such as backpropagation [7] is that the neural nets $f(\boldsymbol{x}; \phi)$ and $g(\boldsymbol{b}; \theta)$, parametrizing the autoencoder, get connected by sampling. The net $f(\boldsymbol{x}; \phi)$ predicts the bit activation probabilities and then a hash $\boldsymbol{b} \in \{0,1\}^B$ is drawn to feed $g(\boldsymbol{b}; \theta)$. Fortunately, one of the main achievements of the last years in deep learning has been adapting backpropagation to "pass" through stochastic layers like these [10]. [3] has shown that this method works well for hashing with Gaussian representations. In the case of discrete distributions the gradients can be estimated using the so-called Gumbel-Softmax reparametrization trick [9]. Experiments in [17] show that this method is stable and effective for hashing.

## E. Semi-Supervised Training

If some examples in the training set $S = \{\boldsymbol{x}^{(\ell)}\}$ have been annotated with class labels describing its semantic content $\boldsymbol{y}^{(\ell)} \subset \mathbb{Y} = \{t_1, t_2, \ldots, t_K\}$, the unsupervised objective can be expanded to exploit this information. Hereafter we assume that the annotations have been one-hot encoded as probability distributions, i.e. $\boldsymbol{y}_j = 1$ if $\boldsymbol{x} \in S$ has been annotated with a label $t_j$ and $\boldsymbol{y}_j = 0$ otherwise. To accommodate the semi-supervised scenario, we assume that only the first $s \ll n$ examples from $S$ have been labelled.

*a) Pointwise Supervision:* A simple way to guide the model towards a more discriminative latent representation for a pattern $\boldsymbol{x}$ is to train the model to learn not only $p(\boldsymbol{x})$ but also $p(\boldsymbol{y}|\boldsymbol{x})$. More specifically, if the neural net implementing the encoder is $f = f_1 \circ \ldots f_{L-1} \circ f_L$, we burden the model with the task of inferring $p(\boldsymbol{y}|\boldsymbol{x})$ from the representation $\boldsymbol{z} = f_1 \circ \ldots \circ f_{L-1}$ computed immediately before the bit-activation probabilities $\alpha(\boldsymbol{x})$. This approach is illustrated in Fig.1. Intuitively, if two patterns $\boldsymbol{x}^{(1)}, \boldsymbol{x}^{(2)}$ have the same annotations, the model should learn that $p(\boldsymbol{y}|\boldsymbol{x}^{(1)}) \approx p(\boldsymbol{y}|\boldsymbol{x}^{(2)})$. As $p(\boldsymbol{y}|\boldsymbol{x})$ is computed from $\boldsymbol{z}$, we should have $p(\boldsymbol{y}|\boldsymbol{x}^{(1)}) \approx p(\boldsymbol{y}|\boldsymbol{x}^{(2)}) \Rightarrow \boldsymbol{z}^{(1)} \approx \boldsymbol{z}^{(2)}$. However, as also the codes are computed from $\boldsymbol{z}$, we should have $\boldsymbol{z}^{(1)} \approx \boldsymbol{z}^{(2)} \Rightarrow \boldsymbol{b}^{(1)} \approx \boldsymbol{b}^{(2)}$. Therefore, the model should learn that patterns with the same annotations have to be allocated in nearby addresses of the hash table.

A similar type of supervision has been used in [3] and [4]. However, in these works, $p(\boldsymbol{y}|\boldsymbol{x})$ is approximated from the representation computed immediately after the decoder's output. Our choice is tailored to the self-supervision method introduced below.

As sketched in Fig.1, we can approximate the distribution $p(\boldsymbol{y}|\boldsymbol{x})$, augmenting the autoencoder with an extra fully connected layer $\hat{\boldsymbol{y}}(\boldsymbol{z}; \psi)$. The parameters $\psi$ of this layer can be jointly trained with the rest of the architecture to minimize the cross-entropy loss between the predicted label distribution for a labelled example $\boldsymbol{x}^{(\ell)}$ and the ground-truth. In the simplest case (one-hot vectors representing mutually exclusive labels), the loss takes the form

$$\mathcal{L}_{\text{sup}}^{(\ell)} = -\mathbb{E}\left[\boldsymbol{y}^{(\ell)} \log p_\psi(\boldsymbol{y}|\boldsymbol{x}^{(\ell)})\right] = -\sum_k \boldsymbol{y}_k^{(\ell)} \log \hat{\boldsymbol{y}}_k^{(\ell)}. \quad (2)$$

*b) Pairwise Supervision:* A more explicit way to enforce a consistency between the similarities in the code space and the similarities in the label space can be obtained by equipping the model with a pairwise loss function. If $\boldsymbol{b}^{(\ell)}$ and $\boldsymbol{b}^{(\ell')}$ denote the codes assigned to a pair of examples $\boldsymbol{x}^{(\ell)}, \boldsymbol{x}^{(\ell')}$ sampled from the labelled dataset, and $\boldsymbol{y}^{(\ell)}, \boldsymbol{y}^{(\ell')}$ denote their ground-truth labels, a loss that penalizes/rewards differences between the codes of similar/dissimilar pairs is

$$\mathcal{L}_{\text{pair}}^{(\ell,\ell')} = I(\boldsymbol{y}^{(\ell)} = \boldsymbol{y}^{(\ell')}) D^+(\boldsymbol{b}^{(\ell)}, \boldsymbol{b}^{(\ell')}) \quad (3)$$
$$- I(\boldsymbol{y}^{(\ell)} \neq \boldsymbol{y}^{(\ell')}) D^-(\boldsymbol{b}^{(\ell)}, \boldsymbol{b}^{(\ell')}),$$

where $D^\pm$ denote distance functions in the Hamming space. Often, $D^+$ is chosen to be the standard Hamming distance $\|\boldsymbol{b}^{(\ell)} - \boldsymbol{b}^{(\ell')}\|_H$, but $D^-$ is shrunk as $D^- = -(\rho - \|\boldsymbol{b}^{(\ell)} - \boldsymbol{b}^{(\ell')}\|_H)_+$ to avoid wasting efforts in separating dissimilar pairs beyond a margin $\rho$. This loss has been used in a plethora of hashing algorithms (see e.g. [16]). In the context of variational autoencoders, it has been proposed in [4].

*c) Self-Supervision:* Both the pointwise and the pairwise schemes of supervision suffer the lack of labelled examples. However, a hashing algorithm using the pairwise loss can deteriorate faster than a method using only pointwise supervision. Arguably, this happens because if the labelled subset is reduced to a fraction $\rho$ of the training set, the fraction of pairs that can be generated reduces to $\rho^2$. This issue makes the method more prone to over-fitting in semi-supervised scenarios with label scarcity. To address this problem, we propose a self-supervised learning mechanism in which the ground-truth labels required for Eqn.(3) and substituted by $\hat{\boldsymbol{y}}(\boldsymbol{z}; \psi)$, the predictions of the pointwisely supervised layer of the autoencoder. To formally define the new loss function, we first write (3) in matrix form. Indeed, as $\boldsymbol{y}^{(\ell)T}\boldsymbol{y}^{(\ell')} = 0$ if the points $\boldsymbol{x}^{(\ell)}, \boldsymbol{x}^{(\ell')}$ have different labels and otherwise $\boldsymbol{y}^{(\ell)T}\boldsymbol{y}^{(\ell')} = 1$, we have that the pairwise loss is equivalent to

$$\mathcal{L}_{\text{pair}}^{(\ell,\ell')} = \boldsymbol{y}^{(\ell)T}\boldsymbol{y}^{(\ell')} D_{\ell,\ell'}^+ - (1 - \boldsymbol{y}^{(\ell)T}\boldsymbol{y}^{(\ell')}) D_{\ell,\ell'}^- , \quad (4)$$

where we have used $D_{\ell,\ell'}^\pm$ as a short-hand for $D^\pm(\boldsymbol{b}^{(\ell)}, \boldsymbol{b}^{(\ell')})$. The self-supervised loss is thus defined as

$$\mathcal{L}_{\text{selfsup}}^{(\ell,\ell')} = \hat{\boldsymbol{y}}^{(\ell)T}\hat{\boldsymbol{y}}^{(\ell')} D_{\ell,\ell'}^+ - (1 - \hat{\boldsymbol{y}}^{(\ell)T}\hat{\boldsymbol{y}}^{(\ell')}) D_{\ell,\ell'}^- . \quad (5)$$

Intuitively, minimizing the pointwise loss (2) requires less annotations than learning a label-consistent hash function. Thus, after some training rounds we will have that $\hat{\boldsymbol{y}}^{(\ell)}$ will approximate $\boldsymbol{y}^{(\ell)}$ for many unlabelled observations. Hereafter the self-supervised loss (5) approximates the more conventional pairwise loss (4). Note in addition that, as the label distribution in (5) is now trainable, the loss can be examined as a function of the labels $\hat{\boldsymbol{y}}$ assigned by the algorithm to the different points of the Hamming space. Reordering the terms,

$$\mathcal{L}_{\text{selfsup}}^{(\ell,\ell')} = (D_{\ell,\ell'}^+ + D_{\ell,\ell'}^-)\hat{\boldsymbol{y}}^{(\ell)T}\hat{\boldsymbol{y}}^{(\ell')} - D_{\ell,\ell'}^- , \quad (6)$$

we see that the loss function penalizes correlations between the label distributions in a way proportional to $D_{\ell,\ell'}^+ + D_{\ell,\ell'}^-$. The loss is 0 if and only if pairs $(\ell, \ell')$ for which $D^\pm > 0$ get assigned orthogonal label distributions. As there is a finite number of (normalized) distributions on $\mathbb{Y}$ which are mutually orthogonal, the proposed loss is minimized by reserving a different labelling to distant regions of the Hamming space ($D^\pm \gg 0$).

### F. Efficient Implementation

The final objective function for training the autoencoder in semi-supervised scenarios is

$$\mathcal{L} = \sum_{\ell=1}^n \mathcal{L}_{\text{unsup}}^{(\ell)} + \beta \sum_{\ell=1}^s \mathcal{L}_{\text{sup}}^{(\ell)} + \alpha \sum_{\ell,\ell'=1}^n \mathcal{L}_{\text{selfsup}}^{(\ell,\ell')} \quad (7)$$

where $\beta, \alpha > 0$ are hyper-parameters. Note that only the supervised loss $\mathcal{L}_{\text{sup}}$ is computed on labelled instances. The unsupervised loss and the self-supervised loss are computed using all the available observations.

---

**Algorithm 1: SSB-VAE.**

**Input:** A set of examples $S = \{\boldsymbol{x}^{(1)}, \ldots, \boldsymbol{x}^{(n)}\}$ and semantic labels $\boldsymbol{y}^{(\ell)}$ for the first $s$.
**Output:** Trained parameters $\phi, \theta, \psi$.

1 Initialize $\phi, \theta, \psi$;
2 **while** *not converged* **do**
3    Randomly split $S$ into $n/M$ batches of size $M$;
4    **foreach** *mini-batch* $B_j$ **do**
5      Predict $\hat{\boldsymbol{y}}^{(\ell)}$ for any $\boldsymbol{x}^{(\ell)} \in B_j$;
6      Average the gradients of (1) and (5) w.r.t. $\phi, \theta, \psi$ among all the examples in $B_j$;
7      Average the gradient of (2) w.r.t. $\psi$ among the labelled examples in $B_j$;
8      Perform backpropagation updates for $\phi, \theta, \psi$;
9    **end foreach**
10 **end while**

---

We optimize (7) using backpropagation. Indeed, thanks to the Gumbel-Softmax estimator [9], we can efficiently compute the gradients of $\mathcal{L}_{\text{unsup}}^{(\ell)}$ and $\mathcal{L}_{\text{selfsup}}^{(\ell,\ell')}$ w.r.t. all the model's parameters $\phi, \theta, \psi$. Being $\hat{\boldsymbol{y}}^{(\ell)}$ independent on the stochastic layer, the gradients of $\mathcal{L}_{\text{sup}}^{(\ell)}$ can be computed classically. However, the direct computation of the total gradient/loss has a quadratic computational complexity in the number of examples. As sketched in Alg.1, we circumvent this problem by forming the pairs required for $\mathcal{L}_{\text{selfsup}}^{(\ell,\ell')}$ at a mini-batch level. In this way, the computational cost of the algorithm is only $\mathcal{O}(nM)$, where $M$ is the mini-batch size, a small constant.

## IV. EXPERIMENTS

We conduct experiments to evaluate the robustness of semi-supervised variational autoencoders for hashing in scenarios of label scarcity. The proposed approach is compared with previous methods on text and image retrieval tasks, widely used to assess this type of algorithms. Our code along with instructions to reproduce the results is made publicly available at: `https://github.com/amacaluso/SSB-VAE`.

*a) Data:* The text retrieval tasks are defined on three annotated corpora: *20 Newsgroups*, containing 18000 newsgroup posts on 20 different topics; *TMC* containing 28000 air traffic reports annotated using 22 tags; and *Google Search Snippets*, with 12000 short documents organized in 8 classes (domains). We define an image retrieval task using the dataset *CIFAR-10*, containing 60000 $32 \times 32$ RGB images of 10 different classes [11]. To facilitate comparisons, we represent the text using TD-IDF features as in [3] and [17]. *20 Newsgroups* (hereafter abbreviated *20News*) and *TMC* are used with the train/validation/test split used in [3]. For *Snippets*, we follow [17] and randomly sample a test set of 1200 texts, a validation set of the same size, and leave the rest for training. Images are represented using deep *VGG* descriptors as in [5]. For *CIFAR-10*, we use the pre-defined test set [11]. A validation set of the same size is randomly sampled from the training set.

*b) Methods:* We compare three semi-supervised methods based on variational autoencoders: (1) **VDHS-S**, a variational autoencoder proposed in [3] employing Gaussian latent variables, unsupervised learning and pointwise supervision; (ii) **PHS-GS**, a variational autoencoder proposed in [4] employing Bernoulli latent variables, unsupervised learning, pointwise supervision and pairwise supervision; and (iii) **SSB-VAE**, our proposed method based on Bernoulli latent variable, unsupervised learning, pointwise supervision and self-supervision. As you could note, other combinations of latent variables and types of supervision are possible. For sake of brevity we compare only with published methods. Note also that, as we use the same features and train/validation/test split that [3] for the text datasets, our results can be directly compared with the performance of many other deep learning methods assessed in this work, as in [4].

*c) Technical Details:* To implement the neural nets corresponding to the encoder/decoder, we adopted the same architectures used in [3], for all the methods. We trained the models using 30 epochs, batch size $M = 100$, and the Adam learning rate scheduler [7]. The KL weight $\lambda$ in Eqn.(1) was set to the values reported in [17]. The parameters $\beta$ and $\alpha$ required for **PHS-GS** and **SSB-VAE** were selected on the validation set, using a logarithmic search grid in the range $[10^{-6}, 10^6]$. For a fair comparison, we also allowed **VDHS-S** to select the weight of the supervised loss in the objective function. The scores reported in figures and tables were obtained as an average over 5 runs. All the codes were implemented in Python 3.7 with TensorFlow 2.1 and executed using a small GPU (GTX 1080Ti).

*d) Evaluation:* To evaluate the effectiveness of the hash codes, each document/image in the test set is used as a query to search for similar items in the training set. Following previous works [4, 5], a relevant search result is one which has the same ground-truth label (topic) as the query. To favour comparisons, the performance is measured using p@100, the precision within the first $k = 100$ retrieved documents/images, sorted according to the Hamming distances of their corresponding hash codes to that of the query. We also compute the mean average precision, the average of p@k varying $k$ from 1 to the length of the retrieved list. This score penalizes missing relevant items among the first positions of the list.

To assess the robustness of the algorithms to label scarcity, we train and evaluate the models at varying levels of supervision $\rho = s/n$, the ratio of labelled examples in the training set. Starting from $\rho = 1$, which represents a fully supervised setting, we stress the algorithms reducing $\rho$ by steps of 0.1 until we get a 10% of supervision.

*e) Results & Discussion:* Table I shows the precision of the different methods on the four datasets used for evaluation. We present results for code lengths of $B = 16$ and $B = 32$ bits. It can be confirmed that when all the training instances are labelled ($\rho = 1$), the model based on pointwise supervision

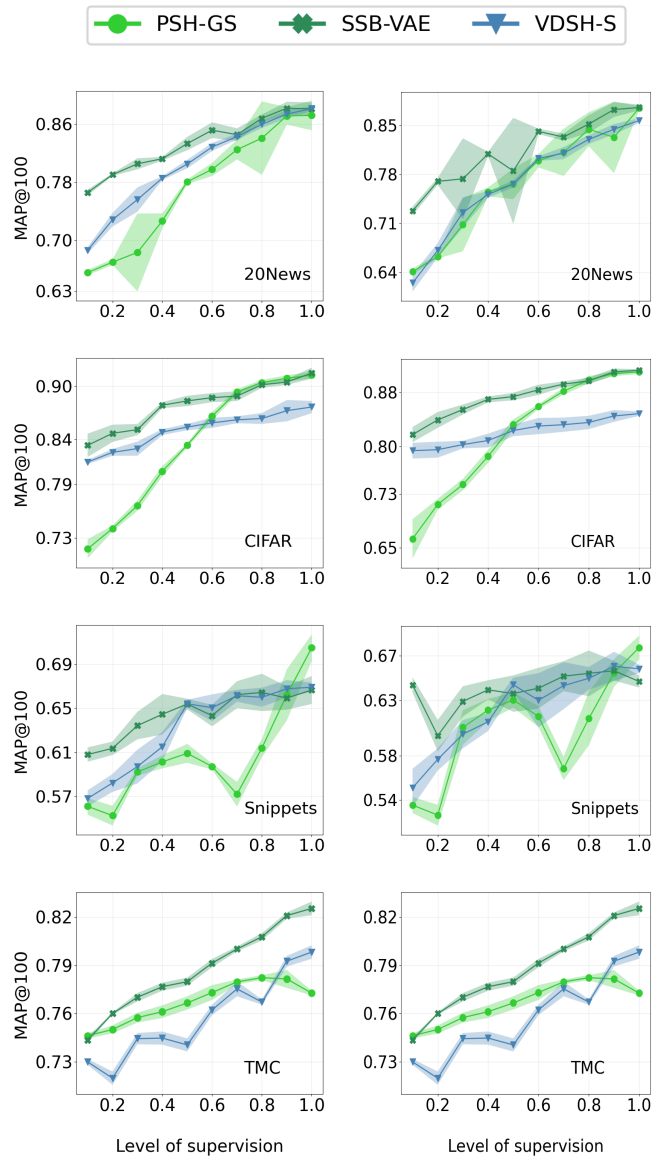

Fig. 2. Mean average precision (MAP@100) of the different algorithms for different levels of supervision. At left, the results with 32 hashing bits. At right, the results with 16 bits.

(PSH-GS) often achieves better results than the method based on pointwise supervision (VDSH-S). This tendency, previously reported in the literature [4], is no longer clear if we reduce the supervision level, specially below 50% ($\rho \leq 0.5$). For instance, with only 20% of training images labeled, VDSH-S gives a precision of 81.6% in CIFAR, 10% over PSH-GS, in absolute terms. We see something similar in Snippets and 20News/32Bits. In these cases, the performance of PSH-GS suffers significantly more the lack of supervision, with a precision loss over 20% in CIFAR, 25% in 20News and 20% in Snippets. To better illustrate this point, we display in Fig.2 the mean average precision (MAP) of the algorithms as a function of $\rho$. In many cases (CIFAR, Snippets and 20News/32Bits) the performance of PSH-GS is clearly decreasing faster as the amount of supervision decreases down to 10%. These results suggest that, in scenarios of label scarcity, learning the label distribution, i.e. minimising the expected pointwise loss, may be easier than learning label consistent hash codes. As the lack of annotations has a quadratic effect on the number of ground-truth pairs that can be used, the pairwise approach is more fragile in semi-supervised scenarios. The results in TMC and 20News/16Bits, in which PSH-GS and VDSH-S deteriorate at more similar rates, may be explained by the fact that PSH-GS implements both types of supervision, pointwise and pairwise.

TABLE I
P@100 OF THE DIFFERENT METHODS FOR DIFFERENT LEVELS OF SUPERVISION $\rho$. A) 32 HASHING BITS AND B) 16 HASHING BITS. THE ALGORITHMS PHS-GS [4] AND VDHS-S [3] ARE ABBREVIATED PSH AND VDSH.

| A) | 20-News | | | CIFAR | | | Snippets | | | TMC | | |
| --- | --- | --- | --- | --- | --- | --- | --- | --- | --- | --- | --- | --- |
| $\rho$ | PSH | SSB-VAE | VDSH | PSH | SSB-VAE | VDSH | PSH | SSB-VAE | VDSH | PSH | SSB-VAE | VDSH |
| 0.1 | 0.589 | **0.734** | 0.648 | 0.687 | **0.825** | 0.805 | 0.501 | **0.565** | 0.540 | 0.738 | **0.750** | 0.730 |
| 0.2 | 0.606 | **0.765** | 0.697 | 0.708 | **0.840** | 0.816 | 0.490 | **0.599** | 0.558 | 0.749 | **0.754** | 0.725 |
| 0.3 | 0.630 | **0.787** | 0.738 | 0.737 | **0.847** | 0.820 | 0.542 | **0.620** | 0.576 | 0.757 | **0.759** | 0.736 |
| 0.4 | 0.682 | **0.791** | 0.771 | 0.781 | **0.873** | 0.838 | 0.551 | **0.620** | 0.595 | 0.765 | **0.775** | 0.740 |
| 0.5 | 0.762 | **0.824** | 0.788 | 0.818 | **0.879** | 0.844 | 0.564 | **0.641** | 0.634 | 0.772 | **0.778** | 0.743 |
| 0.6 | 0.784 | **0.843** | 0.818 | 0.857 | **0.881** | 0.849 | 0.550 | **0.634** | 0.633 | 0.782 | **0.788** | 0.758 |
| 0.7 | 0.815 | **0.841** | 0.831 | **0.889** | 0.880 | 0.852 | 0.553 | 0.644 | **0.648** | 0.790 | **0.795** | 0.768 |
| 0.8 | 0.831 | **0.864** | 0.851 | **0.901** | 0.898 | 0.854 | 0.598 | 0.637 | **0.647** | 0.798 | **0.802** | 0.769 |
| 0.9 | 0.867 | **0.880** | 0.866 | **0.903** | 0.901 | 0.863 | 0.644 | 0.648 | **0.656** | 0.806 | **0.813** | 0.781 |
| 1.0 | 0.866 | **0.878** | 0.876 | 0.906 | **0.910** | 0.867 | **0.696** | 0.657 | 0.661 | 0.806 | **0.818** | 0.788 |

| B) | 20-News | | | CIFAR | | | Snippets | | | TMC | | |
| --- | --- | --- | --- | --- | --- | --- | --- | --- | --- | --- | --- | --- |
| $\rho$ | PSH | SSB-VAE | VDSH | PSH | SSB-VAE | VDSH | PSH | SSB-VAE | VDSH | PSH | SSB-VAE | VDSH |
| 0.1 | 0.595 | **0.711** | 0.582 | 0.635 | **0.816** | 0.781 | 0.482 | **0.621** | 0.522 | 0.723 | **0.725** | 0.705 |
| 0.2 | 0.618 | **0.758** | 0.635 | 0.684 | **0.834** | 0.782 | 0.472 | **0.576** | 0.553 | 0.731 | **0.742** | 0.694 |
| 0.3 | 0.678 | **0.762** | 0.705 | 0.718 | **0.849** | 0.789 | 0.569 | **0.612** | 0.580 | 0.740 | **0.751** | 0.719 |
| 0.4 | 0.731 | **0.799** | 0.733 | 0.765 | **0.866** | 0.796 | 0.589 | **0.627** | 0.591 | 0.743 | **0.759** | 0.721 |
| 0.5 | 0.752 | **0.770** | 0.744 | 0.820 | **0.870** | 0.811 | 0.598 | **0.614** | 0.623 | 0.750 | **0.763** | 0.715 |
| 0.6 | 0.789 | **0.833** | 0.789 | 0.851 | **0.879** | 0.817 | 0.588 | **0.629** | 0.613 | 0.756 | **0.775** | 0.739 |
| 0.7 | 0.802 | **0.829** | 0.794 | 0.877 | **0.884** | 0.818 | 0.551 | **0.634** | 0.627 | 0.764 | **0.782** | 0.753 |
| 0.8 | 0.837 | **0.848** | 0.815 | **0.894** | 0.893 | 0.821 | 0.596 | 0.629 | **0.636** | 0.768 | **0.790** | 0.744 |
| 0.9 | 0.826 | **0.870** | 0.831 | 0.904 | **0.906** | 0.832 | 0.635 | **0.647** | 0.647 | 0.768 | **0.803** | 0.770 |
| 1.0 | 0.872 | **0.873** | 0.846 | 0.906 | **0.909** | 0.836 | **0.666** | 0.641 | 0.649 | 0.759 | **0.808** | 0.777 |

In Tab.I and Fig.2 we can see that the proposed method, which uses the ground-truth labels to learn the label distribution, but employs its own predictions to implement the pairwise loss, is much more robust to the lack of annotations. Its performance decreases more smoothly as the fraction of labelled instances reduces down, achieving noticeable improvements on PSH-GS for small amounts of supervision. This illustrates the interest of the approach for semi-supervised scenarios. For instance, with only 10% of training documents labeled, the proposed method gives a precision of 73.4% in 20News/32Bits, an (absolute) improvement of 14.5% compared to PSH-GS and 8.6% over VDSH-S. For the same level of supervision, it gets a precision of 81.6% in CIFAR/32Bits, a clear improvement over the 63.5% of the pairwise approach. In Snippets/16Bits, SSB-VAE provides a precision only 2% less than the precision achieved in the fully supervised case. In some cases (TMC), the pairwise approach is able to keep its advantage on VDSH-S almost uniformly as the supervision becomes lower. If this is the case, the proposed method is still competitive or better than the best baseline. Although the most significant improvements are obtained for smaller $\rho$, we also confirm that in scenarios of label abundance, using pairwise supervision based on the ground-truth distributions does not give a very significant advantage over the proposed approach. Indeed, in many cases (6/8) SSB-VAE achieves slightly better scores. To obtain an overall conclusion regarding the robustness of the proposed method, two types of statistical tests are conducted. We employ the Friedman test to assess whether there is enough statistical evidence to reject the hypothesis that the three methods are statistically equivalent (in terms of $p@100$), when considering different levels of supervision. In this design, the method (SSB-VAE, PSH, VDSH) serves as the group variable, and the level of supervision serves as the blocking variable. In addition, when rejecting the null hypothesis of Friedman's test, we compare the proposed method against PSH and VDSH using the Nemenyi post-hoc test, to check for statistically significant differences. The obtained p-values are presented in Tab.II. In all but two cases we obtain values below 5%.

## V. Conclusions

We studied the performance of semi-supervised hashing algorithms based on variational autoencoders in scenarios of label scarcity. It was found that training the model to explicitly preserve pairwise similarities derived from the annotations, often yields better results than using pointwise supervision (only), confirming results of previous works. However, we also found that methods based on this type of supervision tend to deteriorate more sharply when the number of labelled observations decreases. To overcome this problem, we proposed a new type of supervision in which the model uses its own beliefs about the class distribution to enforce a consistency between the similarities in the code space and the similarities in the label space. Experiments in text and image retrieval tasks confirmed that this method degrades much more gracefully when the models are stressed with scarcely annotated data, and very often outperforms the baselines by a significant margin. As in scenarios of label abundance, the proposed method proved to be competitive or better than the best baseline, we can conclude that it is a robust approach to semi-supervised hashing. In future work we plan to equip the method with adaptive loss weights and extend the experiments to cross-domain information retrieval.

TABLE II
P-VALUES OF THE STATISTICAL TESTS

|  |  | Friedman Test | Nemenyi Test | |
|---|---|---|---|---|
|  |  |  | PSH | VDSH |
| 32 Bits | 20-NEWS | $1.1 \times 10^{-4}$ | $6.3 \times 10^{-5}$ | $3.7 \times 10^{-2}$ |
|  | SNIPPETS | $7.4 \times 10^{-3}$ | $1.0 \times 10^{-2}$ | $8.9 \times 10^{-1}$ |
|  | TMC | $4.5 \times 10^{-5}$ | $6.5 \times 10^{-2}$ | $2.3 \times 10^{-5}$ |
|  | CIFAR | $2.0 \times 10^{-2}$ | $1.1 \times 10^{-1}$ | $1.2 \times 10^{-2}$ |
| 16 Bits | 20-NEWS | $5.0 \times 10^{-4}$ | $4.9 \times 10^{-3}$ | $1.0 \times 10^{-3}$ |
|  | SNIPPETS | $7.4 \times 10^{-3}$ | $1.0 \times 10^{-2}$ | $8.9 \times 10^{-1}$ |
|  | TMC | $2.2 \times 10^{-4}$ | $1.9 \times 10^{-2}$ | $1.6 \times 10^{-4}$ |
|  | CIFAR | $1.8 \times 10^{-3}$ | $1.9 \times 10^{-2}$ | $2.2 \times 10^{-3}$ |

REFERENCES

[1] R. Baeza-Yates and B. Ribeiro-Neto. *Modern Information Retrieval*. ACM, 1999.
[2] M. A. Carreira-Perpinán and R. Raziperchikolaei. "Hashing with binary autoencoders". *Proc. CVPR*. 2015, pp. 557–566.
[3] S. Chaidaroon and Y. Fang. "Variational deep semantic hashing for text documents". *SIGIR*. 2017, pp. 75–84.
[4] S. Z. Dadaneh et al. "Pairwise Supervised Hashing with Bernoulli Variational Auto-Encoder and Self-Control Gradient Estimator". *Proc. UAI*. 2020.
[5] T.-T. Do et al. "Learning to hash with binary deep neural network". *Proc. ECML*. 2016, pp. 219–234.
[6] Y. Gong and S. Lazebnik. "Iterative quantization: A procrustean approach to learning binary codes". *Proc. CVPR*. 2011, pp. 817–824.
[7] I. Goodfellow et al. *Deep learning*. MIT press, 2016.
[8] P. Indyk and R. Motwani. "Approximate nearest neighbors: towards removing the curse of dimensionality". *Proc. ACM STOC*. 1998, pp. 604–613.
[9] E. Jang et al. "Categorical Reparameterization with Gumbel-softmax". *Proc. ICLR*. 2017.
[10] D. P. Kingma and M. Welling. "Auto-encoding variational Bayes". *Proc. ICLR*. 2014.
[11] A. Krizhevsky et al. "Imagenet classification with deep convolutional neural networks". *NIPS*. 2012, pp. 1097–1105.
[12] B. Kulis and K. Grauman. "Kernelized Locality-sensitive Hashing". *IEEE Pattern Anal. Mach. Intell.* 34.6 (2012), pp. 1092–1104.
[13] H. Lai et al. "Simultaneous feature learning and hash coding with deep neural networks". *Proc. CVPR*. 2015, pp. 3270–3278.
[14] C. Li et al. "Self-supervised adversarial hashing networks for cross-modal retrieval". *Proc. CVPR*. 2018, pp. 4242–4251.
[15] H. Liu et al. "Deep supervised hashing for fast image retrieval". *Proc. CVPR*. 2016, pp. 2064–2072.
[16] J. Lu et al. "Deep hashing for scalable image search". *IEEE Trans. Image Process.* 26.5 (2017), pp. 2352–2367.
[17] F. Mena and R. Ñanculef. "A binary variational autoencoder for hashing". *Proc. CIARP*. 2019, pp. 131–141.
[18] Y. Pan et al. "Semi-supervised hashing with semantic confidence for large scale visual search". *Proc. SIGIR*. 2015, pp. 53–62.
[19] R. Salakhutdinov and G. Hinton. "Semantic hashing". *International Journal of Approximate Reasoning* 50.7 (2009), pp. 969–978.
[20] T. Song et al. "Semi-supervised manifold-embedded hashing with joint feature representation and classifier learning". *Pattern Recognition* 68 (2017), pp. 99–110.
[21] C. Strecha et al. "LDAHash: Improved matching with smaller descriptors". *IEEE Pattern Anal. Mach. Intell.* 34.1 (2012), pp. 66–78.
[22] I. Triguero et al. "Self-labeled techniques for semi-supervised learning: taxonomy, software and empirical study". *Knowledge and Information Systems* 42.2 (2015), pp. 245–284.
[23] J. Wang et al. "Semi-supervised hashing for large-scale search". *IEEE Pattern Anal. Mach. Intell.* 34.12 (2012), pp. 2393–2406.
[24] Q. Wang et al. "Semantic hashing using tags and topic modeling". *Proc. SIGIR*. ACM. 2013, pp. 213–222.
[25] Y. Weiss et al. "Spectral hashing". *NIPS*. 2009.
[26] G. Wu et al. "Joint image-text hashing for fast large-scale cross-media retrieval using self-supervised deep learning". *IEEE Transactions on Industrial Electronics* 66.12 (2018), pp. 9868–9877.
[27] H. Yang et al. "Adaptive labeling for hash code learning via neural networks". *Proc. ICIP*. 2019, pp. 2244–2248.
[28] D. Zhang et al. "Self-taught hashing for fast similarity search". *Proc. SIGIR*. 2010, pp. 18–25.
[29] H. Zhang et al. "Play and rewind: Optimizing binary representations of videos by self-supervised temporal hashing". *ACM Multimedia*. 2016, pp. 781–790.